\documentclass[letterpaper,journal]{IEEEtran}
\usepackage{amsmath,amsfonts,amssymb,graphicx,cite,booktabs,array,xcolor,url,placeins,dblfloatfix}
\usepackage[caption=false,font=footnotesize]{subfig}
\definecolor{TOMASgreen}{RGB}{0,120,52}
\newcommand{\best}[1]{\textbf{#1}}
\newcommand{\second}[1]{\textcolor{TOMASgreen}{\textbf{#1}}}

\usepackage{graphicx}

\newcommand{\biophoto}[1]{%
    \includegraphics[width=1in,height=1.25in]{#1}%
}

\newcommand{\biospace}{%
    \vspace{-3\baselineskip}%
}

\usepackage{tikz}
\usetikzlibrary{calc}

\newcommand{\ForecastComparison}[1]{%
\begin{tikzpicture}
    \node[anchor=south west, inner sep=0] (img) at (0,0) {%
        \includegraphics[width=.80\textwidth]{#1}%
    };

    \node[anchor=south, font=\footnotesize\bfseries]
        at ($(img.north west)!0.1667!(img.north east)+(0,4pt)$)
        {TOMAS (Ours)};

    \node[anchor=south, font=\footnotesize\bfseries]
        at ($(img.north west)!0.5!(img.north east)+(0,4pt)$)
        {USST};

    \node[anchor=south, font=\footnotesize\bfseries]
        at ($(img.north west)!0.8333!(img.north east)+(0,4pt)$)
        {Uni-Hand};

    \node[anchor=east, align=center,
          font=\footnotesize, xshift=-5pt]
        at ($(img.south west)!0.8333!(img.north west)$)
        {EgoPAT3D-DT\\Seen};

    \node[anchor=east, align=center,
          font=\footnotesize, xshift=-5pt]
        at ($(img.south west)!0.5!(img.north west)$)
        {EgoPAT3D-DT\\Unseen};

    \node[anchor=east, align=center,
          font=\footnotesize, xshift=-5pt]
        at ($(img.south west)!0.1667!(img.north west)$)
        {H2O-PT};
\end{tikzpicture}%
}

\begin{document}
\title{TOMAS: Topology-Aware Operation Modeling and Adaptive Staging for Egocentric Hand Forecasting}
\author{Guanzhong Sun, Junyi Ma, Yuxuan Wu, Erhang Zhang, Chenyang Xu, Yanzi Miao, and Hesheng Wang\textsuperscript{*}%
\thanks{Guanzhong Sun and Yanzi Miao are with the School of Information and Control Engineering, China University of Mining and Technology, Xuzhou, China.}%
\thanks{Junyi Ma, Yuxuan Wu, Erhang Zhang, Chenyang Xu, and Hesheng Wang are with the School of Automation and Intelligent Sensing, Shanghai Jiao Tong University, Shanghai, China.}%
\thanks{Yuxuan Wu is also with the Shanghai Innovation Institute, Shanghai, China.}%
\thanks{Hesheng Wang is also with the Shanghai Key Laboratory of Navigation and Location Based Services, Shanghai, China.}}
% TODO before submission: verify all authors' institutional e-mail addresses and add funding acknowledgments if applicable.
\markboth{IEEE Transactions on Circuits and Systems for Video Technology}%
{Sun \MakeLowercase{\textit{et al.}}: TOMAS}
\maketitle
\begin{abstract}
Forecasting future hand movements from egocentric videos plays an important role in understanding human intentions and hand-object interactions. However, different manipulation intentions can lead to substantially different hand trajectories even when interacting with the same object. Existing methods have not sufficiently explored the intrinsic relationships between articulated hand movements and object semantics to capture such operation-dependent motion patterns. Therefore, we propose TOMAS, a topology-aware operation modeling and adaptive staging framework for egocentric hand forecasting. Specifically, we develop a topology-aware hand-object reasoning mechanism that associates hand skeletal structures and articulation dynamics with object semantics and spatial relationships to jointly infer potential interaction targets and latent manipulation operations. The resulting object--operation representations provide interaction conditions for future hand trajectory prediction, with supervision adapted to the annotations available in each dataset. Furthermore, we introduce an adaptive staging mechanism that organizes future hand movements into ordered latent stages with learnable temporal boundaries. Stage-specific geometric subgoals are then incorporated to guide trajectory refinement according to the underlying manipulation patterns. 
Experimental results on multiple publicly available benchmarks demonstrate that TOMAS achieves state-of-the-art performance in multi-dimensional and multi-target hand motion forecasting.

\end{abstract}

\section{Introduction}
\label{sec:introduction}

Forecasting future hand trajectories from egocentric video is important for embodied learning and human-centered interaction \cite{liu2022oct,bao2023usst}. Existing methods have advanced object-centered interaction modeling \cite{liu2022oct}, motion-aware temporal dynamics \cite{ma2025madiff}, and general hand-motion representations \cite{ma2026unihand}. While these advances improve future motion representation and generation, the actor's intended manipulation remains largely implicit in the forecasting process.

\begin{figure}[t]
\centering
\includegraphics[width=\columnwidth]{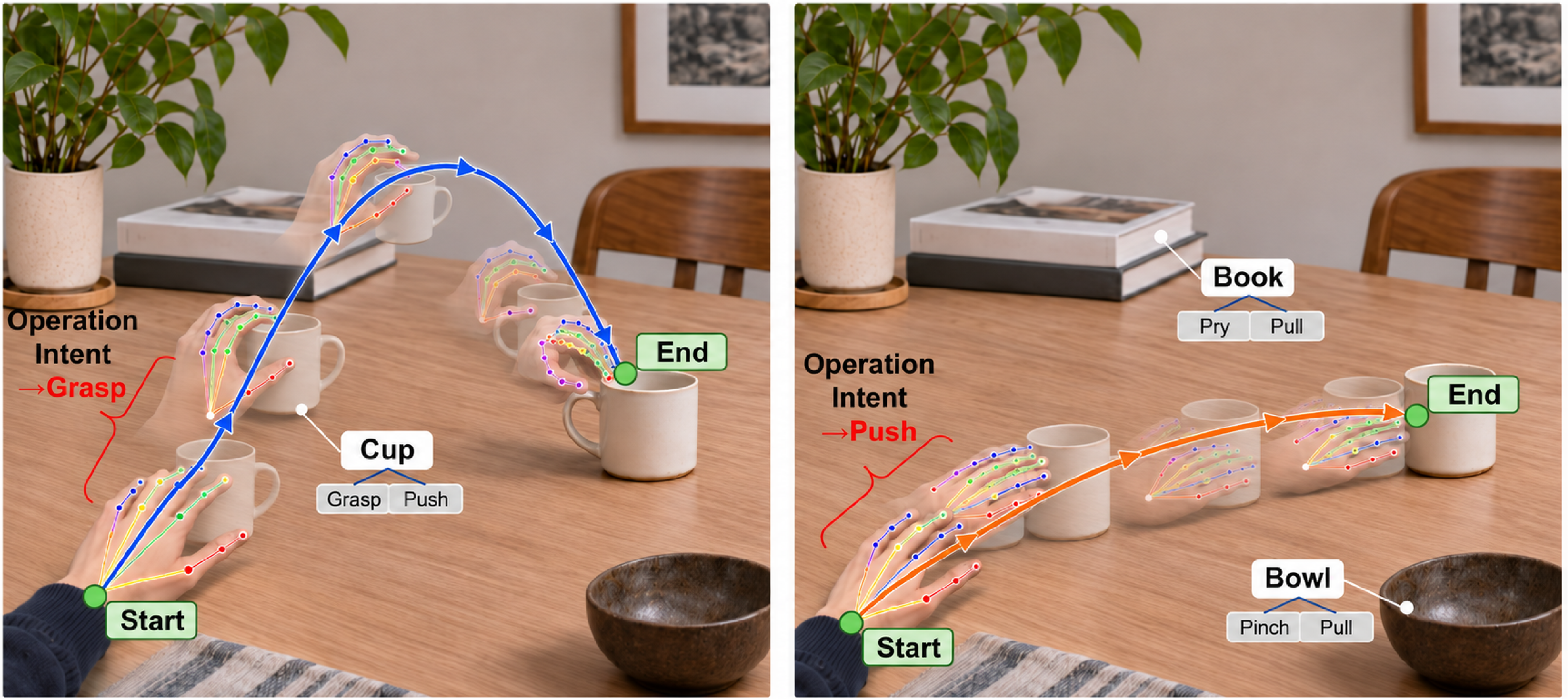}
\caption{\textbf{Operation-dependent hand forecasting.} The same object can support different operations with distinct future hand trajectories. }
\label{fig:motivation}
\vspace{-0.5cm}
\end{figure}

Observed motion does not uniquely determine where a hand will move next. Even with accurate observations of past hand motion and nearby objects, the intended action may remain ambiguous before contact. As illustrated in Fig.~\ref{fig:motivation}, the same object can support different operations, such as grasping, pushing, and pulling, each involving distinct hand configurations and future motion paths. Moreover, even when the interaction target is known, different operations unfold over different temporal intervals and impose different intermediate spatial constraints. For example, a grasp may require a preparatory hand adjustment before contact, while a push can involve a more direct approach followed by sustained displacement. This motivates a central question in egocentric hand forecasting: How can a model jointly reason about candidate objects and latent operations from past observations, and use the resulting interaction representations to guide the temporal and geometric structure of future hand trajectories?

This question presents three connected challenges. The first is \emph{interaction ambiguity}. Hand motion and object appearance must be considered jointly to distinguish plausible targets and operations. Resolving this ambiguity leads to the second challenge, \emph{temporal evolution}. An operation is not a single motion state but a progression whose phases and durations depend on the interaction context. The third challenge is \emph{spatial realization}. Different phases require intermediate geometric objectives, so even trajectories with similar endpoints may follow substantially different paths. These challenges call for a model that connects operation reasoning, temporal organization, and geometric trajectory generation.

We propose TOMAS, a Topology-Aware Operation Modeling and Adaptive Staging framework for egocentric hand forecasting. TOMAS first constructs object-operation hypotheses from observed hand articulation, skeletal topology, candidate-object context, and hand-object relations. Rather than conditioning prediction on a single global motion representation, the model uses these hypotheses to capture operation-dependent alternatives. TOMAS then organizes future motion into ordered latent stages conditioned on the inferred interaction context. Learned stage boundaries accommodate nonuniform temporal evolution without requiring stage annotations. Each stage also predicts a geometric subgoal that anchors the corresponding part of the future path. These stage-conditioned subgoals guide a hierarchical trajectory decoder, linking the intended interaction to both the timing and geometry of predicted hand motion.

The main contributions are summarized as follows.
\begin{itemize}

\item We propose TOMAS, a novel topology-aware operation
reasoning framework for egocentric hand forecasting.
By jointly exploiting articulated hand dynamics,
skeletal topology, object semantics, and spatial
interaction relationships, TOMAS models the dependencies
between candidate interaction targets and latent
manipulation operations. 
The resulting object--operation representations
enable interaction-conditioned trajectory forecasting,
with auxiliary target and operation supervision used
when corresponding annotations are available.

\item We introduce a manipulation-conditioned adaptive
staging mechanism to capture the temporal organization
of future hand movements. By learning ordered latent
stages with sample-adaptive temporal boundaries, our
method characterizes the evolving motion patterns
associated with different manipulations.

\item We develop a stage-wise geometric subgoal-guided
forecasting strategy that connects latent motion
stages with intermediate spatial objectives, providing
structured trajectory refinement beyond endpoint constraints.

\item We conduct comprehensive experiments based on the
existing evaluation metrics to demonstrate
that TOMAS predicts comparably reasonable hand trajectories compared to state-of-the-art baselines.
\end{itemize}

\section{Related Work}
\label{sec:related_work}

\subsection{Egocentric Hand Motion Forecasting}

Egocentric hand forecasting aims to infer future hand movements from partial first-person observations. Early approaches established interaction context as an important source of predictive information. FHOI models forthcoming hand behavior through motor attention \cite{liu2020fhoi}, and OCT jointly predicts hand trajectories and object interaction hotspots \cite{liu2022oct}. USST introduces uncertainty-aware state transitions for future motion \cite{bao2023usst}, while VRB learns contact locations and subsequent wrist trajectories as transferable interaction affordances \cite{bahl2023vrb}. These studies connect anticipated motion with the surrounding scene and possible contact regions. Complementary action-anticipation work studies interaction regions \cite{roy2024interaction} and verb--noun uncertainty \cite{guo2024decoupling}.
Recent methods have advanced plausible trajectory generation. Diff-IP2D formulates hand-object interaction prediction through diffusion \cite{ma2025diffip2d}. MADiff introduces motion-aware state-space modeling that incorporates camera motion and visual-language cues \cite{ma2025madiff}. Uni-Hand unifies forecasting across multiple input modalities, spatial representations, and prediction targets \cite{ma2026unihand}. HOIMotion investigates the role of egocentric object geometry in forecasting human motion \cite{hu2024hoimotion}. Together, these developments provide powerful representations of motion dynamics and scene context. TOMAS advances interaction-conditioned forecasting by reasoning over candidate object-operation hypotheses and relating them to the temporal and geometric organization of future hand motion.

\subsection{Hand-Object Interaction and Operation Reasoning}

Manipulation depends on how articulated hands interact with objects, rather than object location alone. Atomic hand--object interaction modeling has also related hand and object cues to task references \cite{fan2022atomic}. H2O and HOI4D provide benchmarks for understanding first-person hand-object activities \cite{kwon2021h2o,liu2022hoi4d}. Related egocentric perception methods study relation modeling for action recognition \cite{li2023egorelation} and wearer--object relations for text-supervised segmentation \cite{shi2025ctdn}. HOnnotate, ARCTIC, and HOT3D support detailed modeling of hand articulation, object pose, and interaction geometry \cite{hampali2020honnotate,fan2023arctic,banerjee2025hot3d}. Hand reconstruction methods such as HaMeR further enable the recovery of articulated hand configurations from images \cite{pavlakos2024hamer}.
Beyond pose and geometry, manipulation understanding requires cues about what the hand is doing to an object. Joint hand-pose and action modeling relates articulation patterns to activity \cite{wen2023handaction}, while contact-state recognition and object-state learning capture complementary interaction dynamics \cite{shiota2024contactstate,xue2024objectstate}. Fine-grained hand-object dynamics provide another perspective on operation-dependent changes \cite{pei2025handobject}. These lines of research motivate a representation in which hand topology, evolving articulation, candidate-object semantics, and relative geometry are considered together. TOMAS uses these cues to form object-operation hypotheses that condition trajectory generation, allowing different plausible manipulations of the same object to induce distinct motion predictions.

\subsection{Temporal Structure and Geometric Motion Generation}

Hand manipulation also has an internal temporal organization. Its successive phases can differ in duration, movement direction, and spatial objective. Temporal action modeling provides relevant tools for describing such structure. Related egocentric action anticipation methods model observed-to-future video segments \cite{guan2023mtrace}, fuse visual and semantic features \cite{cao2024vstransgru}, and account for prediction uncertainty through multiple experts \cite{qi2025antmoe}. MS-TCN learns progressively refined temporal representations \cite{farha2019mstcn}, and ASFormer models long-range dependencies with attention \cite{yi2021asformer}. DiffAct introduces diffusion-based temporal segmentation \cite{liu2023diffact}, while DiffAnt extends generative modeling to future action anticipation \cite{zhong2023diffant}. In motion generation, conditional diffusion models show how structured information can guide plausible continuous motion \cite{tevet2023mdm}. TOMAS connects temporal organization directly to interaction-conditioned hand forecasting. It represents the prediction horizon with ordered latent stages whose boundaries depend on the observed interaction context. Stage-specific operation representations and geometric subgoals provide complementary conditions for hierarchical trajectory generation. This formulation brings together the potential manipulation, its temporal progression, and the intermediate spatial constraints of the future hand path within a unified forecasting framework.

\section{Proposed Method}
\label{sec:method}

\subsection{Problem Formulation and Framework}
\label{sec:formulation}

\begin{figure*}[!t]
\centering
\includegraphics[width=.8\textwidth]{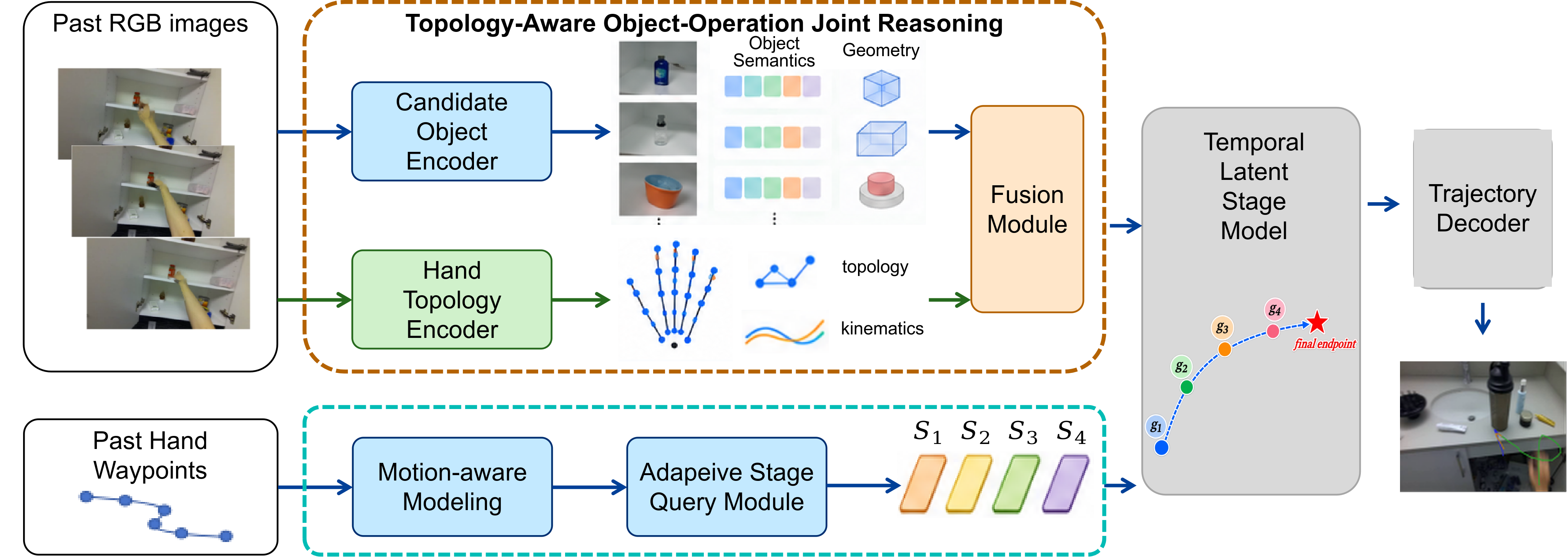}
\vspace{-0.25cm}
\caption{System overview of TOMAS. Given observed egocentric video and historical hand trajectories, TOMAS employs topology-aware hand--object reasoning to jointly model candidate-object semantics and latent operation representations. The resulting interaction-aware representations guide ordered latent motion stages and stage-wise geometric subgoals for future hand trajectory forecasting.}
\label{fig:overview}
\vspace{-0.5cm}
\end{figure*}

Given an observed egocentric video sequence $\mathcal I^p$ and historical hand-center trajectory $\mathcal H^p$, we construct an observation context $X$ comprising articulated hand joints $\mathcal J^p$, camera-motion information $\mathcal E^p$, and candidate-object tracks $\mathcal O^p$. These cues are obtained from available annotations or observation-based estimation and preprocessing, according to the dataset protocol. TOMAS jointly models hand articulation, candidate-object semantics, and latent operations to learn interaction-aware representations. These representations guide ordered latent motion stages and geometric subgoals to forecast the future hand-center trajectory $\widehat{\mathcal H}^{f}=\{\widehat{\mathbf h}_t\}_{t=1}^{T_f}$ in $d\in\{2,3\}$ dimensions. Figure~\ref{fig:overview} illustrates the framework.

\subsection{Topology-Aware Object--Operation Joint Reasoning}
\label{sec:interaction}

\begin{figure}[!t]
\centering
\includegraphics[width=\columnwidth]{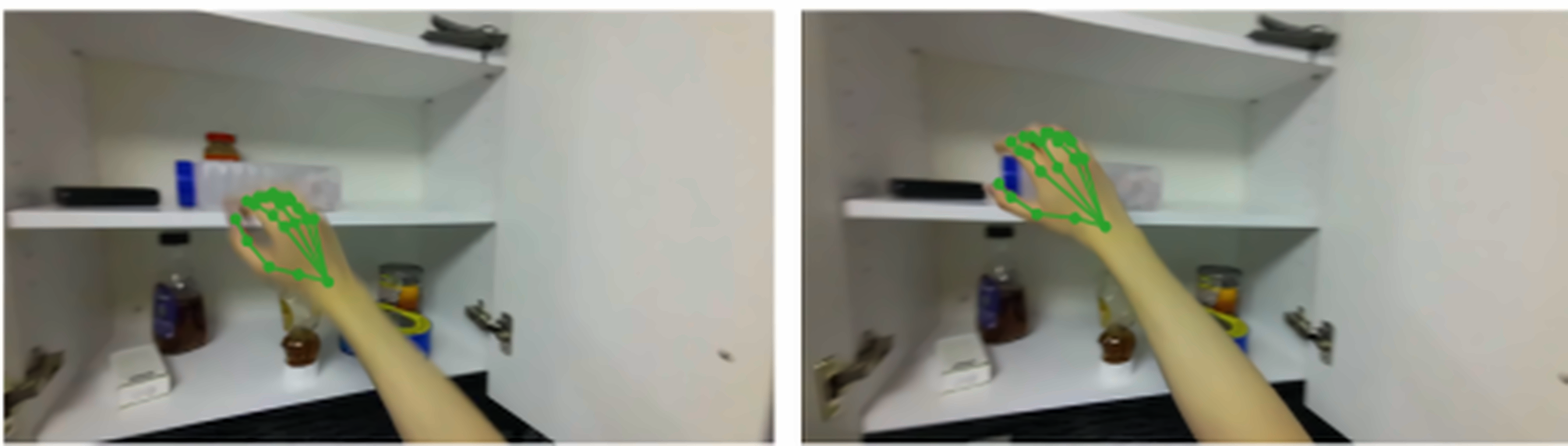}
\vspace{-0.5cm}
\caption{Visualization of articulated hand motion and skeletal topology.}
\label{fig:joint_observation}
\vspace{-0.5cm}
\end{figure}

\textbf{Articulation-aware observations.}
Depending on the dataset, we obtain 21 hand joints from official H2O annotations or ViTPose-based estimation. These joints form an anatomical graph $\mathcal G=(\mathcal V,\mathcal E_{\mathcal G})$, whose edges encode skeletal connections. For each joint $j$ at observed time $t$, we encode root-relative position, temporal displacement, bone displacement, and temporal bone change. Graph propagation captures anatomical dependencies, while temporal aggregation models evolving finger configurations independently of hand-center motion. Figure~\ref{fig:joint_observation} illustrates two observed sequences with distinct articulation dynamics.

\textbf{Object-relative interaction context.} Object proposals and observed tracks are encoded with appearance semantics, bounding-box geometry, segmentation masks, and motion cues. Candidate semantics use CLIP features \cite{radford2021clip}, object proposals use Mask R-CNN \cite{he2017maskrcnn}, and frame-level RGB context uses ConvNeXt \cite{liu2022convnext}. Each candidate $o_i$ is paired with hand-relative spatial and temporal features to form an object token $\mathbf z_i^{\rm obj}$ and a relation token $\mathbf z_i^{\rm rel}$. We organize the observed hand center, topology, egomotion, and RGB streams as a memory $\mathbf M$. Rather than applying the same scene summary to every object, TOMAS makes each candidate query the parts of history relevant to its own interaction context:
\begin{equation}
\begin{split}
 \widetilde{\mathbf z}_i^{\rm int}&=\mathbf q_i+\operatorname{CrossAttn}(\mathbf q_i,\mathbf M,\mathbf M),\\
 \mathbf q_i&=E_o(\mathbf z_i^{\rm obj},\mathbf z_i^{\rm rel}).
\end{split}
\label{eq:candidate_memory}
\end{equation}
A learnable null option models the absence of a suitable observed candidate. Candidate-specific queries can thus emphasize different temporal and modality cues even within the same scene.

\textbf{Joint object--operation reasoning.} Jointly reasoning about an object and a possible manipulation is central to TOMAS. For a valid candidate $o_i$ and a learned operation component $a_k$, the two heads estimate
\begin{equation}
\begin{gathered}
 p_i=P(o_i\mid X),\\
 q_{k\mid i}=P(a_k\mid o_i,X),\\
 P(o_i,a_k\mid X)=p_i\,q_{k\mid i}.
\end{gathered}
\label{eq:joint}
\end{equation}
The highest-scoring object--operation pairs are encoded as interaction-conditioning tokens $\mathbf c_{ik}$, each defining a distinct condition for future trajectory generation. The factorized distribution represents both competition among candidate targets and multiple latent operations associated with the same object. 

\begin{figure*}[!t]
\centering
\includegraphics[width=.75\textwidth]{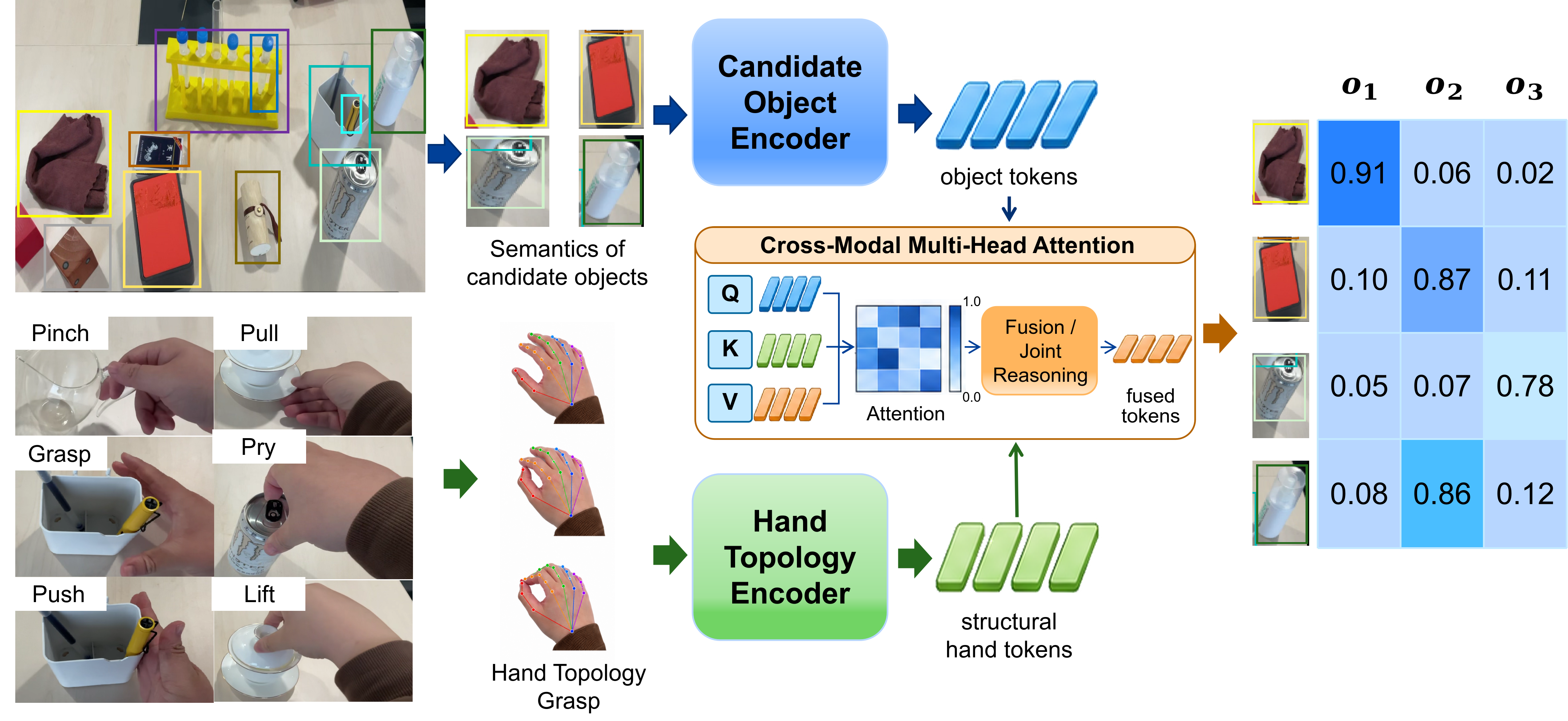}
\vspace{-0.35cm}
\caption{Topology-aware object--operation joint reasoning. Observed object candidates and articulated hand configurations are encoded as candidate-specific interaction representations. Cross-modal memory retrieval supports joint object--operation scoring and the formation of interaction-conditioning tokens.}
\label{fig:operation_reasoning}
\vspace{-0.3cm}
\end{figure*}

The object-target distribution $p_i$, including the null option, is evaluated in Sec.~\ref{sec:target}, while its training supervision is detailed in Sec.~\ref{sec:training}. Figure~\ref{fig:operation_reasoning} illustrates candidate-conditioned joint reasoning, and Fig.~\ref{fig:memory_attention} visualizes candidate-specific retrieval patterns.

\subsection{Interaction-Conditioned Adaptive Motion Stages}
\label{sec:stage}
A future hand path reflects not only which interaction is plausible but also how the corresponding motion evolves. For each retained object--operation pair $(i,k)$, TOMAS uses the interaction-conditioning token $\mathbf c_{ik}$ to initialize future motion features and incorporate observed kinematics and egomotion. A motion-aware Mamba-based temporal backbone provides long-range state propagation \cite{ma2025madiff}; its role here is to supply interaction-conditioned motion features $\{\mathbf f_{ik,t}\}$ from which the temporal stage structure is inferred. Consequently, object--operation evidence acts before the trajectory is decoded and continues to shape the future-time representation.

\begin{figure*}[!t]
\centering
\includegraphics[width=.75\textwidth]{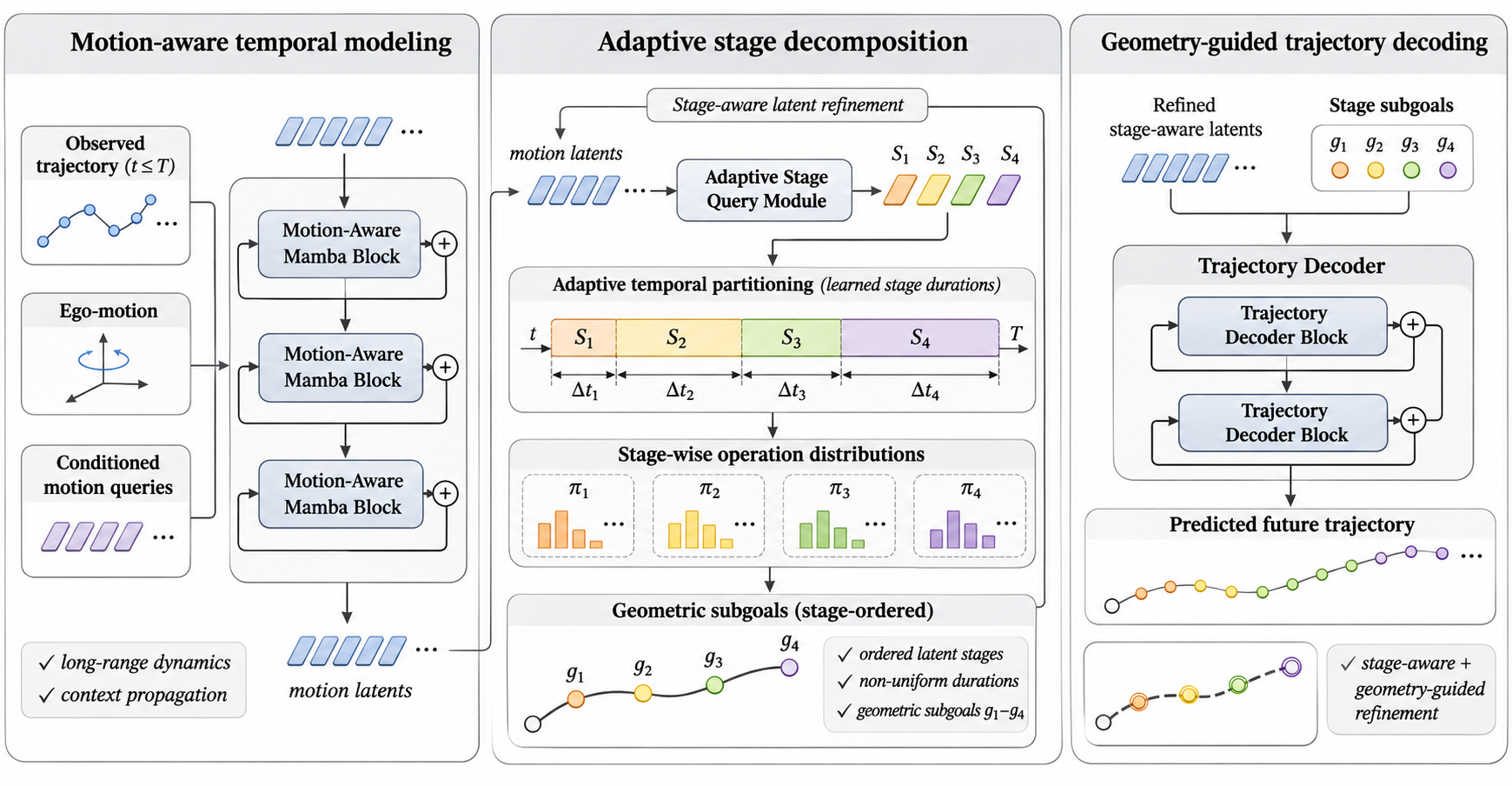}
\vspace{-0.35cm}
\caption{Interaction-conditioned adaptive staging and geometry-guided trajectory generation. Interaction-conditioned motion features determine ordered stage allocations and stage-level operation mixtures. The resulting candidate-relative geometric subgoals guide continuous trajectory decoding and refinement.}
\label{fig:stage}
\vspace{-0.5cm}
\end{figure*}

\textbf{Ordered temporal allocation.} Let $K_s$ denote the number of latent motion stages. Conditioned stage queries predict positive duration scores $\rho_s^{ik}>0$ that are normalized into stage fractions. Their cumulative sums form ordered transition positions:
\begin{equation}
 \Delta_s^{ik}=\frac{\rho_s^{ik}}{\sum_{r=1}^{K_s}\rho_r^{ik}},\qquad
 \beta_s^{ik}=\sum_{r=1}^{s}\Delta_r^{ik}.
 \label{eq:boundaries}
\end{equation}
The ordered boundaries define differentiable stage weights $\alpha_{t,s}^{ik}\geq0$, with $\sum_s\alpha_{t,s}^{ik}=1$, allowing neighboring stages to contribute smoothly around transitions. The future-time representation becomes
\begin{equation}
 \mathbf z_{t,ik}^{\rm stage}=
 \sum_{s=1}^{K_s}\alpha_{t,s}^{ik}\mathbf z_s^{ik}.
 \label{eq:stage_mix}
\end{equation}
Stage durations and motion characteristics are therefore jointly conditioned on the object--operation representation. Stage order is a learned temporal organization, not an imposed sequence. Unlike independent temporal attention weights, the cumulative parameterization explicitly preserves the order of transitions. Figure~\ref{fig:stage} gives an architectural view of this interaction-conditioned staging process.

\subsection{Stage-Wise Geometric Subgoals and Trajectory Generation}
\label{sec:subgoal}
A latent operation can imply different intermediate hand configurations even when candidate targets and final hand positions are similar. TOMAS expresses this distinction through stage-level operation distributions $\boldsymbol\pi_s^{ik}$ and candidate-relative geometric subgoals. For an observed candidate anchor $\mathbf a_i$, the stage subgoal is
\begin{equation}
 \mathbf g_s^{ik}=\mathbf a_i+
 G_{\rm sub}(\mathbf z_s^{ik},\boldsymbol\pi_s^{ik},
 \mathbf z_i^{\rm rel},\mathbf c_{ik}).
 \label{eq:subgoal}
\end{equation}
The operation index $a_k$ specifies the latent operation component of the selected object--operation pair, whereas $\boldsymbol\pi_s^{ik}$ describes the evolving mixture of latent operation components within each stage. Together, they connect pair-level interaction context with stage-level motion variation.

Conditioned residual refinement integrates these geometric preferences over the predicted horizon:
\begin{equation}
 \widetilde{\mathbf f}_{ik,t}=\mathbf f_{ik,t}+
 \sum_{s=1}^{K_s}\alpha_{t,s}^{ik}
 R_s\!\left(\mathbf f_{ik,t},\mathbf g_s^{ik},
 \boldsymbol\pi_s^{ik},\mathbf c_{ik}\right).
 \label{eq:refine}
\end{equation}
A hierarchical decoder converts the refined feature stream into a residual over a kinematic trajectory initialization $\mathbf h_t^{\rm init}$:
\begin{equation}
 \widehat{\mathbf h}_{ik,t}=\mathbf h_t^{\rm init}+
 D_{\rm traj}(\widetilde{\mathbf f}_{ik,t}).
 \label{eq:decode}
\end{equation}
This operation-to-geometry pathway allows the inferred object--operation relationship to shape both the temporal progression and intermediate geometry of the generated path. As illustrated in Fig.~\ref{fig:stage}, ordered stage representations and candidate-relative subgoals jointly condition hierarchical trajectory generation. The subgoals provide continuous geometric guidance instead of prescribing waypoints through which the hand must pass. Section~\ref{sec:temporal_analysis} examines how these conditions affect predicted motion.

\subsection{Training and Inference}
\label{sec:training}
Training combines the trajectory reconstruction term, late-horizon refinement, observed-motion-based egomotion prediction, and endpoint supervision. The motion objective is
\begin{equation}
 \mathcal L_{\rm motion}=
 \mathcal L_{\rm traj}+\lambda_{\rm tail}\mathcal L_{\rm tail}
 +\lambda_{\rm ego}\mathcal L_{\rm ego}
 +\lambda_{\rm end}\mathcal L_{\rm end}.
 \label{eq:loss}
\end{equation}

Here $\mathcal L_{\rm end}$ measures the final-position error of the trajectory selected from the object--operation pairs.
The training objective is
$\mathcal L=\mathcal L_{\rm motion}+
\lambda_{\rm intent}\mathcal L_{\rm intent}$.
For H2O-PT, training-set action annotations
provide auxiliary supervision for the target
and verb distributions, with
$\lambda_{\rm intent}=0.1$.
Target supervision is defined over the
observed object candidate indices, while
verb supervision uses annotated
operation categories.
For EgoPAT3D-DT and EGTEA Gaze+,
$\lambda_{\rm intent}=0$, and selection
probabilities are optimized through
probability-weighted trajectory losses
and differentiable decoder conditioning.

\section{Experiments}
\label{sec:experiments}

\subsection{Datasets, Evaluation Protocol, and Implementation}

We evaluate TOMAS on EgoPAT3D-DT \cite{li2022egopat3d}, H2O-PT \cite{kwon2021h2o,bao2023usst}, and EGTEA Gaze+ \cite{li2023egtea}. EgoPAT3D-DT and H2O-PT support 2D and 3D hand trajectory forecasting, with EgoPAT3D-DT providing seen and unseen evaluation splits. For these two benchmarks, the observed and future portions occupy 60\% and 40\% of each forecasting window, respectively, with main trajectory windows of up to 40 frames. We report mean average displacement error (ADE) and final displacement error (FDE) to evaluate trajectory accuracy over the prediction horizon and at the final timestep. Three-dimensional errors are measured in meters, while two-dimensional errors are computed in normalized image coordinates.

The articulated-hand inputs follow the corresponding dataset protocols. H2O-PT uses official annotated 3D hand joints and their 2D projections for the respective forecasting settings. EgoPAT3D-DT and EGTEA Gaze+ use estimated 2D hand joints obtained through the Uni-Hand/ViTPose pipeline \cite{ma2026unihand,xu2022vitpose}. For EgoPAT3D-DT, these 2D joint features are also used as historical articulation inputs in the 3D forecasting setting.

The main TOMAS models are trained for 350 epochs with three random seeds, and Full-model results are averaged across seeds. EgoPAT3D-DT and H2O-PT use the best validation checkpoint per seed, whereas EGTEA Gaze+ uses the final epoch-350 checkpoints. Ablations A1--A9 use a single seed. For EgoPAT3D-DT and H2O-PT, we adopt published baseline results and protocols from Uni-Hand \cite{ma2026unihand}. For EGTEA Gaze+, we adapt representative baselines, including Uni-Hand, to a unified 1.5-second observation and 0.5-second prediction protocol, training them for 350 epochs and evaluating their final checkpoints.
The 3D and 2D quantitative comparisons on EgoPAT3D-DT and H2O-PT are reported in Tables~\ref{tab:main_3d} and~\ref{tab:main_2d}, respectively.

\begin{table*}[!t]
\centering
\caption{3D hand trajectory forecasting on EgoPAT3D-DT and H2O-PT. Best and second-best scores are in bold and green, respectively.}
\vspace{-0.25cm}
\label{tab:main_3d}
\footnotesize
\setlength{\tabcolsep}{4pt}
\renewcommand{\arraystretch}{1.04}
\begin{tabular*}{\textwidth}{@{\extracolsep{\fill}}lcccccc@{}}
\toprule
& \multicolumn{2}{c}{EgoPAT3D-DT (seen)} & \multicolumn{2}{c}{EgoPAT3D-DT (unseen)} & \multicolumn{2}{c}{H2O-PT}\\
\cmidrule(lr){2-3}\cmidrule(lr){4-5}\cmidrule(l){6-7}
Method & ADE$\downarrow$ & FDE$\downarrow$ & ADE$\downarrow$ & FDE$\downarrow$ & ADE$\downarrow$ & FDE$\downarrow$\\
\midrule
CVH~\cite{ma2025diffip2d} & 1.100 & 1.278 & 0.952 & 1.018 & 0.146 & 0.148 \\
OCT~\cite{liu2022oct} & 0.370 & 0.524 & 0.309 & 0.397 & 0.103 & 0.126 \\
USST~\cite{bao2023usst} & 0.183 & 0.341 & 0.120 & \second{0.185} & 0.031 & \second{0.052} \\
S-Mamba~\cite{wang2025smamba} & 0.185 & 0.355 & 0.138 & 0.207 & 0.038 & 0.074 \\
Diff-IP3D~\cite{ma2025diffip2d} & 0.199 & 0.377 & 0.156 & 0.229 & 0.049 & 0.081 \\
EMAG3D~\cite{hatano2024emag} & 0.207 & 0.365 & 0.166 & 0.209 & 0.062 & 0.085 \\
EER3D~\cite{ma2025eer} & 0.195 & 0.377 & 0.158 & 0.233 & 0.047 & 0.075 \\
MADiff3D~\cite{ma2025madiff} & 0.183 & 0.363 & 0.139 & 0.224 & 0.032 & 0.059 \\
Uni-Hand~\cite{ma2026unihand} & \second{0.170} & \second{0.336} & \second{0.118} & 0.189 & \second{0.030} & \best{0.050} \\
\midrule
TOMAS (ours) & \best{0.158} & \best{0.324} & \best{0.103} & \best{0.174} & \best{0.025} & \best{0.050} \\
\bottomrule
\end{tabular*}
\vspace{-0.25cm}
\end{table*}

\begin{table*}[!t]
\centering
\caption{2D hand trajectory forecasting on EgoPAT3D-DT and H2O-PT. Best and second-best scores are in bold and green, respectively.}
\vspace{-0.25cm}
\label{tab:main_2d}
\footnotesize
\setlength{\tabcolsep}{4pt}
\renewcommand{\arraystretch}{1.04}
\begin{tabular*}{\textwidth}{@{\extracolsep{\fill}}lcccccc@{}}
\toprule
& \multicolumn{2}{c}{EgoPAT3D-DT (seen)} & \multicolumn{2}{c}{EgoPAT3D-DT (unseen)} & \multicolumn{2}{c}{H2O-PT}\\
\cmidrule(lr){2-3}\cmidrule(lr){4-5}\cmidrule(l){6-7}
Method & ADE$\downarrow$ & FDE$\downarrow$ & ADE$\downarrow$ & FDE$\downarrow$ & ADE$\downarrow$ & FDE$\downarrow$\\
\midrule
CVH~\cite{ma2025diffip2d} & 0.180 & 0.230 & 0.188 & 0.221 & 0.206 & 0.208 \\
OCT~\cite{liu2022oct} & 0.108 & 0.122 & 0.091 & 0.147 & 0.387 & 0.381 \\
USST~\cite{bao2023usst} & 0.082 & 0.118 & 0.060 & 0.087 & 0.040 & \second{0.068} \\
S-Mamba~\cite{wang2025smamba} & 0.084 & 0.141 & 0.071 & 0.118 & 0.051 & 0.094 \\
Diff-IP2D~\cite{ma2025diffip2d} & 0.080 & 0.130 & 0.066 & 0.087 & 0.042 & 0.074 \\
EMAG~\cite{hatano2024emag} & 0.095 & 0.143 & 0.092 & 0.139 & 0.054 & 0.093 \\
EER~\cite{ma2025eer} & 0.072 & 0.121 & 0.056 & \second{0.080} & 0.039 & 0.069 \\
MADiff~\cite{ma2025madiff} & 0.065 & \second{0.105} & 0.054 & 0.086 & 0.039 & \second{0.068} \\
Uni-Hand~\cite{ma2026unihand} & \second{0.064} & 0.109 & \second{0.052} & 0.087 & \second{0.036} & \best{0.065} \\
\midrule
TOMAS (ours) & \best{0.053} & \best{0.092} & \best{0.042} & \best{0.075} & \best{0.035} & 0.069 \\
\bottomrule
\end{tabular*}
\vspace{-0.25cm}
\end{table*}

As reported in Tables~\ref{tab:main_3d} and~\ref{tab:main_2d}, TOMAS achieves the best performance in all six 3D metrics, including a tie with Uni-Hand on H2O-PT FDE. For 2D forecasting, TOMAS obtains the lowest errors in five of the six metrics, with consistent improvements on both the seen and unseen EgoPAT3D-DT splits. On H2O-PT, TOMAS achieves the lowest 2D ADE, while Uni-Hand retains a lower FDE. These results demonstrate the competitive performance of TOMAS across different forecasting dimensions and interaction scenarios.

As shown in Table~\ref{tab:main_eg}, TOMAS achieves the lowest ADE of $0.140$ on EGTEA Gaze+ under the dataset-specific short-horizon protocol, compared with $0.181$ for Uni-Hand. Together, the results across the three benchmarks support the effectiveness of interaction-conditioned trajectory forecasting under different spatial and observation settings.

\begin{table}[!t]
\centering
\caption{Short-horizon comparison on EGTEA Gaze+.}
\vspace{-0.25cm}
\label{tab:main_eg}
\footnotesize\setlength{\tabcolsep}{8pt}
\begin{tabular}{lcc}
\toprule
Method & ADE$\downarrow$ & FDE$\downarrow$ \\
\midrule
OCT~\cite{liu2022oct} & 0.160 & \best{0.160} \\
USST~\cite{bao2023usst} & 0.200 & 0.233 \\
Uni-Hand~\cite{ma2026unihand} & 0.181 & 0.208 \\
\midrule
TOMAS (ours) & \best{0.140} & \second{0.165} \\
\bottomrule
\end{tabular}
\vspace{-0.5cm}
\end{table}

\subsection{Qualitative Forecasting on 2D and 3D Benchmarks}
\label{sec:qualitative}

\begin{figure*}[!t]
\centering
\ForecastComparison{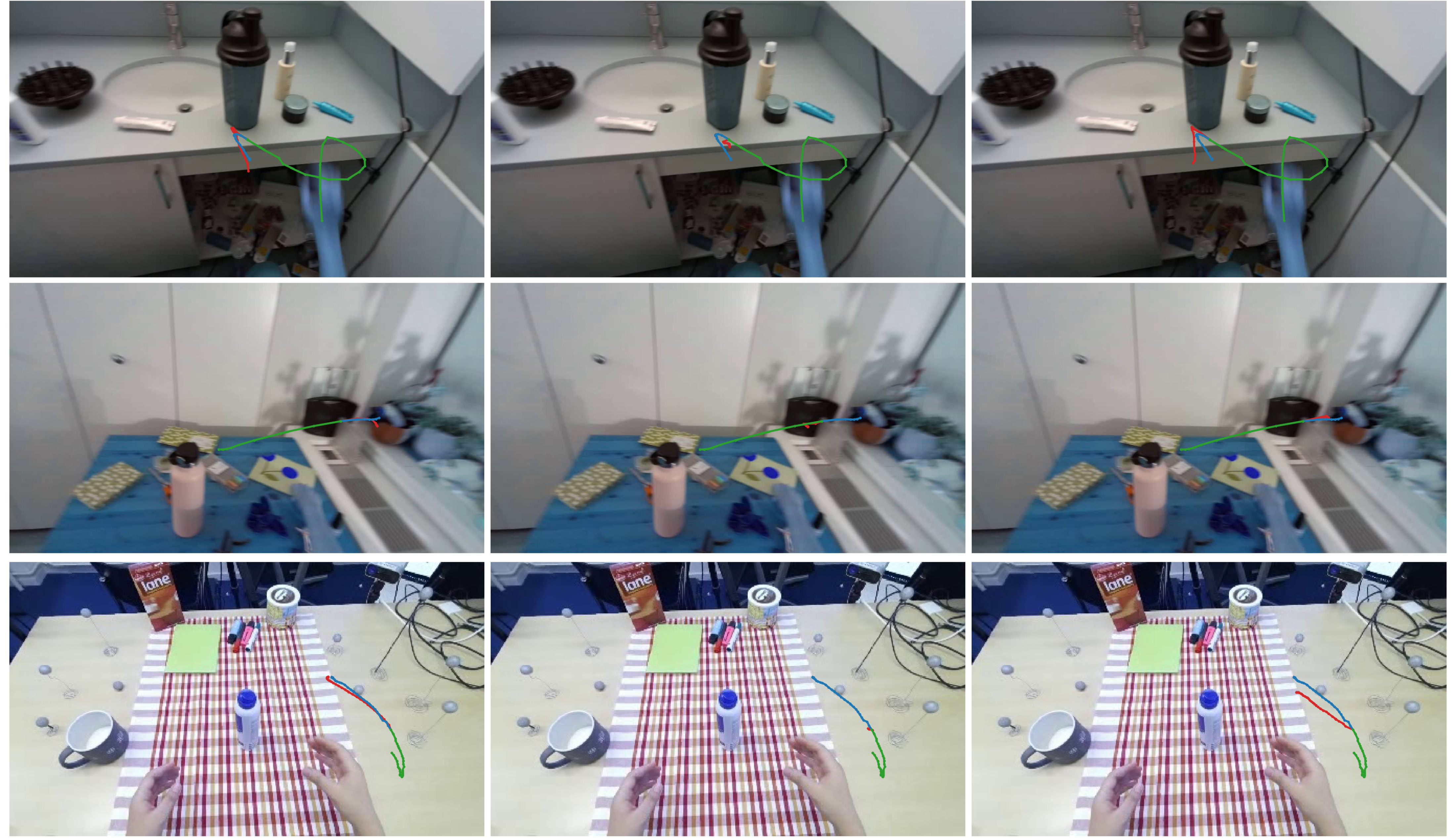}
\caption{Qualitative comparison of 2D hand-center forecasting on EgoPAT3D-DT and H2O-PT. Green, blue, and red denote observed trajectories, future ground truth, and predicted trajectories, respectively.}
\label{fig:qual}
\vspace{-0.5cm}
\end{figure*}

\begin{figure*}[!t]
\centering
\ForecastComparison{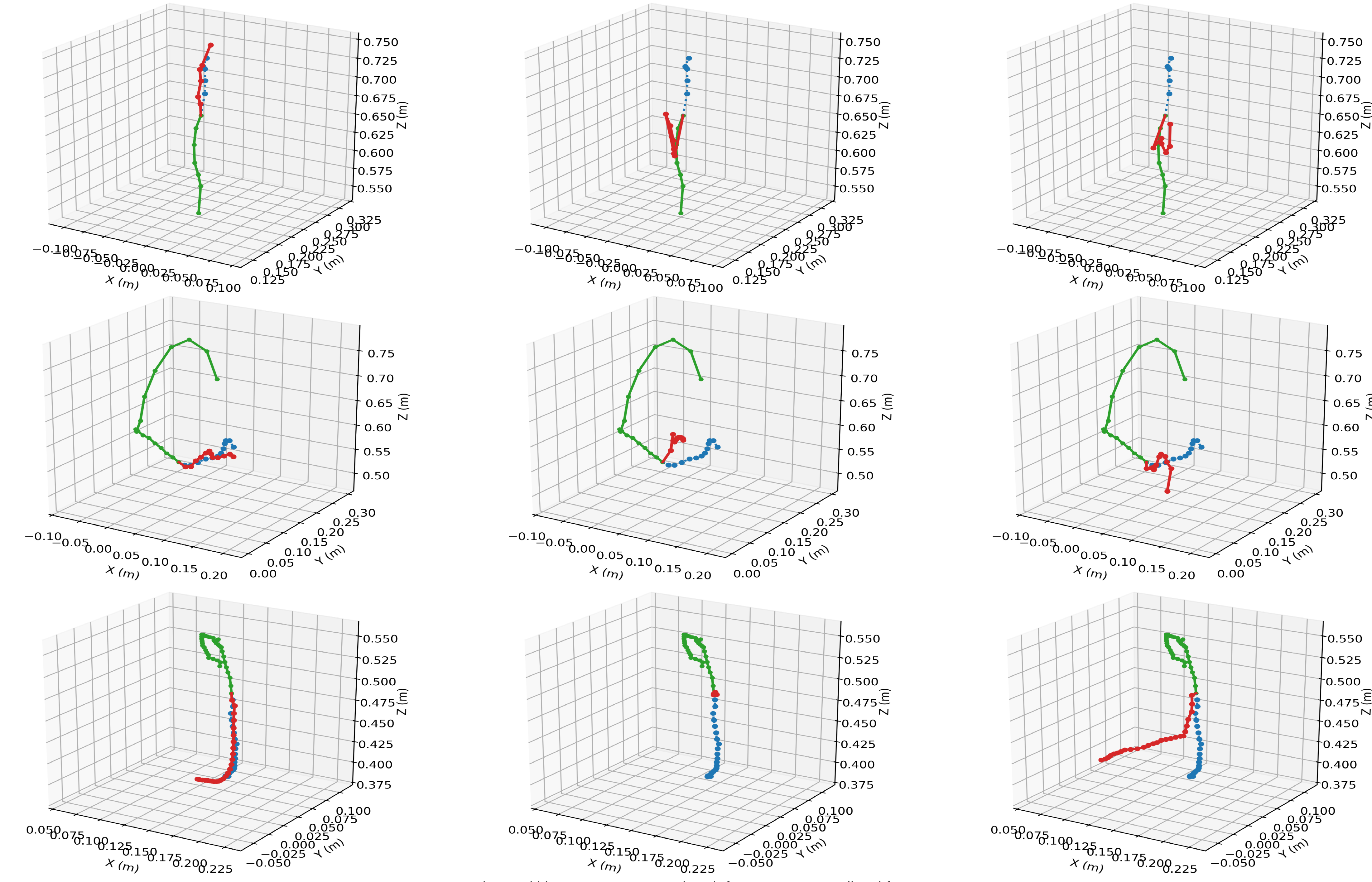}
\caption{Qualitative comparison of 3D hand-center
forecasting on EgoPAT3D-DT and H2O-PT.
Green, blue, and red denote observed trajectories,
future ground truth, and predicted trajectories,
respectively. Coordinates are measured in meters.}
\label{fig:qual3d}
\vspace{-0.5cm}
\end{figure*}

To complement the quantitative comparisons in Tables~\ref{tab:main_2d} and~\ref{tab:main_3d}, Figs.~\ref{fig:qual} and~\ref{fig:qual3d} present qualitative 2D and 3D forecasting results, respectively. We compare TOMAS with USST and Uni-Hand on the seen and unseen EgoPAT3D-DT splits and H2O-PT, visualizing the observed trajectories, future ground truth, and predicted hand motion.

The examples show how TOMAS follows the evolving geometry of future hand motion while approaching its final destination. This trajectory-level behavior is consistent with the improvements in ADE across the 2D and 3D benchmarks in Tables~\ref{tab:main_3d} and~\ref{tab:main_2d}.

\subsection{Analysis of Interaction and Motion Mechanisms}
\label{sec:ablation}

We investigate nine component ablations (A1--A9) to examine the roles of interaction reasoning and motion generation in TOMAS. The Full-model results are averaged over three seeds, whereas each ablation variant reports a single-seed result. The comparisons therefore indicate performance trends associated with individual components rather than seed-matched estimates. Tables~\ref{tab:ablation_3d} and~\ref{tab:ablation_2d} report the 3D and 2D results on EgoPAT3D-DT and H2O-PT, respectively, while Table~\ref{tab:ablation_eg} presents the results on EGTEA Gaze+. A1--A3 examine hand articulation, skeletal topology, and relative hand--object geometry; A4--A5 investigate joint object--operation reasoning and global interaction conditioning; A6--A8 evaluate ordered stage boundaries, geometric subgoals, and stage-level operation conditioning; and A9 removes endpoint supervision.

\textbf{Global and stage-level interaction conditioning.}

A5 and A8 yield higher FDE than Full across all seven evaluation settings. On EGTEA Gaze+, their FDE values are $0.175$ and $0.173$, respectively, compared with $0.165$ for Full. These results support the complementary roles of global interaction conditioning and stage-level operation cues.

\textbf{Hand articulation, temporal organization, and geometric guidance.}
Removing hand-joint information (A1) produces the largest degradation on EGTEA Gaze+, increasing ADE/FDE from $0.140/0.165$ to $0.159/0.185$. A6 replaces ordered stage boundaries with unconstrained softmax temporal attention, while A7 removes geometric subgoals. 
Both A6 and A7 yield higher FDE than
the Full-model reference across all seven
evaluation settings. Overall, the ablations support two complementary aspects of TOMAS: topology-aware object--operation reasoning provides interaction information for future motion prediction, while ordered motion stages and geometric subgoals help organize its temporal and spatial evolution. The local analysis in Fig.~\ref{fig:geom_refinement} further illustrates the effect of geometric guidance.

\begin{table*}[!t]
\centering
\caption{Ablation results for 3D hand trajectory forecasting on EgoPAT3D-DT and H2O-PT. Full reports the three-seed average, while A1--A9 report single-seed results.}
\vspace{-0.25cm}
\label{tab:ablation_3d}
\footnotesize
\setlength{\tabcolsep}{4.0pt}
\renewcommand{\arraystretch}{1.05}
\begin{tabular*}{\textwidth}{@{\extracolsep{\fill}}lcccccc@{}}
\toprule
& \multicolumn{2}{c}{EgoPAT3D-DT (seen)} & \multicolumn{2}{c}{EgoPAT3D-DT (unseen)} & \multicolumn{2}{c}{H2O-PT}\\
\cmidrule(lr){2-3}\cmidrule(lr){4-5}\cmidrule(l){6-7}
Variant & ADE$\downarrow$ & FDE$\downarrow$ & ADE$\downarrow$ & FDE$\downarrow$ & ADE$\downarrow$ & FDE$\downarrow$\\
\midrule
Full (TOMAS) & 0.158 & 0.324 & 0.103 & 0.174 & 0.025 & 0.050 \\
\midrule
A1 (w/o Hand Joints) & 0.161 & 0.326 & 0.106 & 0.175 & 0.027 & 0.051 \\
A2 (w/o Graph Prop.) & 0.161 & 0.329 & 0.105 & 0.182 & 0.025 & 0.051 \\
A3 (w/o Rel. Geometry) & 0.160 & 0.326 & 0.103 & 0.178 & 0.025 & 0.051 \\
A4 (Decoupled Obj.-Op.) & 0.161 & 0.326 & 0.107 & 0.175 & 0.025 & 0.051 \\
A5 (w/o Global Cond.) & 0.163 & 0.329 & 0.109 & 0.178 & 0.026 & 0.053 \\
A6 (Softmax Stage Attn.) & 0.161 & 0.326 & 0.106 & 0.175 & 0.025 & 0.051 \\
A7 (w/o Geom. Subgoals) & 0.161 & 0.326 & 0.106 & 0.175 & 0.025 & 0.051 \\
A8 (w/o Stage Op. Cond.) & 0.160 & 0.327 & 0.103 & 0.179 & 0.025 & 0.051 \\
A9 (w/o Endpoint Sup.) & 0.161 & 0.327 & 0.105 & 0.179 & 0.025 & 0.053 \\
\bottomrule
\end{tabular*}
\vspace{-0.25cm}
\end{table*}

\begin{table*}[!t]
\centering
\caption{Ablation results for 2D hand trajectory forecasting on EgoPAT3D-DT and H2O-PT. Full reports the three-seed average, while A1--A9 report single-seed results.}
\vspace{-0.25cm}
\label{tab:ablation_2d}
\footnotesize
\setlength{\tabcolsep}{4.0pt}
\renewcommand{\arraystretch}{1.05}
\begin{tabular*}{\textwidth}{@{\extracolsep{\fill}}lcccccc@{}}
\toprule
& \multicolumn{2}{c}{EgoPAT3D-DT (seen)} & \multicolumn{2}{c}{EgoPAT3D-DT (unseen)} & \multicolumn{2}{c}{H2O-PT}\\
\cmidrule(lr){2-3}\cmidrule(lr){4-5}\cmidrule(l){6-7}
Variant & ADE$\downarrow$ & FDE$\downarrow$ & ADE$\downarrow$ & FDE$\downarrow$ & ADE$\downarrow$ & FDE$\downarrow$\\
\midrule
Full (TOMAS) & 0.053 & 0.092 & 0.042 & 0.075 & 0.035 & 0.069 \\
\midrule
A1 (w/o Hand Joints) & 0.056 & 0.094 & 0.044 & 0.080 & 0.037 & 0.072 \\
A2 (w/o Graph Prop.) & 0.054 & 0.094 & 0.043 & 0.077 & 0.036 & 0.072 \\
A3 (w/o Rel. Geometry) & 0.054 & 0.093 & 0.044 & 0.077 & 0.036 & 0.071 \\
A4 (Decoupled Obj.-Op.) & 0.054 & 0.094 & 0.043 & 0.077 & 0.035 & 0.069 \\
A5 (w/o Global Cond.) & 0.055 & 0.094 & 0.044 & 0.080 & 0.037 & 0.073 \\
A6 (Softmax Stage Attn.) & 0.054 & 0.094 & 0.043 & 0.077 & 0.035 & 0.070 \\
A7 (w/o Geom. Subgoals) & 0.054 & 0.093 & 0.044 & 0.076 & 0.035 & 0.070 \\
A8 (w/o Stage Op. Cond.) & 0.054 & 0.093 & 0.044 & 0.077 & 0.036 & 0.071 \\
A9 (w/o Endpoint Sup.) & 0.054 & 0.094 & 0.044 & 0.078 & 0.036 & 0.071 \\
\bottomrule
\end{tabular*}
\vspace{-0.5cm}
\end{table*}

\begin{table}[!t]
\centering
\caption{Ablation results for short-horizon hand trajectory forecasting on EGTEA Gaze+. Full reports the three-seed average, while A1--A9 report single-seed results.}
\vspace{-0.25cm}
\label{tab:ablation_eg}
\footnotesize
\setlength{\tabcolsep}{11pt}
\renewcommand{\arraystretch}{1.10}
\begin{tabular}{lcc}
\toprule
Variant & ADE$\downarrow$ & FDE$\downarrow$\\
\midrule
Full (TOMAS) & 0.140 & 0.165\\
\midrule
A1 (w/o Hand Joints) & 0.159 & 0.185\\
A2 (w/o Graph Prop.) & 0.145 & 0.173\\
A3 (w/o Rel. Geometry) & 0.145 & 0.173\\
A4 (Decoupled Obj.-Op.) & 0.145 & 0.173\\
A5 (w/o Global Cond.) & 0.147 & 0.175\\
A6 (Softmax Stage Attn.) & 0.145 & 0.173\\
A7 (w/o Geom. Subgoals) & 0.145 & 0.173\\
A8 (w/o Stage Op. Cond.) & 0.145 & 0.173\\
A9 (w/o Endpoint Sup.) & 0.145 & 0.174\\
\bottomrule
\end{tabular}
\vspace{-0.5cm}
\end{table}

\subsection{Native Target-Object Ranking}
\label{sec:target}

We further evaluate the target-object ranking capability of TOMAS on the H2O-PT validation set, as summarized in Table~\ref{tab:target}. This analysis examines whether topology-aware interaction reasoning helps identify operation-related objects from historical observations. Each 2D or 3D setting contains 1,735 hand windows, including 1,608 with derived operation targets. Each window follows the 60\% observation protocol, spans up to 64 frames, and includes up to eight observed object candidates and a null option.

For this validation-set diagnostic, evaluation targets are derived from validation action annotations, separately from the training-set annotations used for auxiliary target and verb supervision on H2O-PT. We select an ongoing action extending into the prediction interval or, otherwise, the earliest future non-background action. Its associated object is matched to an observed candidate using mean projected-center distance with a fixed threshold of $0.25$. The validation-derived target indices are used to evaluate the predicted object probabilities and are not provided as model inputs. These geometrically associated scene-action targets are not independently verified hand-specific interaction annotations.

For target-bearing windows $\mathcal S_{\rm int}$, the native object distribution $p_i(o)$ defines
\begin{equation}
 \begin{aligned}
 \hat o_i&=\arg\max_{o\in\mathcal C_i\cup\{\varnothing\}}p_i(o),\\
 \mathrm{Acc}_{\rm obj}&=\frac{1}{N_{\rm int}}
 \sum_{i\in\mathcal S_{\rm int}}\mathbf 1[\hat o_i=o_i^*],
 \end{aligned}
 \label{eq:target_acc}
\end{equation}

All 1,608 action-target windows were matched to observed candidates, yielding 100\% association coverage and identical overall and conditional accuracy. This coverage does not establish manually verified detector recall. We use seed-0 checkpoints selected by validation mADE, making this a validation-set diagnostic.

\begin{table}[!t]
\centering
\caption{Native H2O-PT target ranking. Each model has 1,735 windows; 1,608 have derived interaction targets. Coverage is 100\% under the candidate association rule.}
\vspace{-0.25cm}
\label{tab:target}
\footnotesize\setlength{\tabcolsep}{3.0pt}\renewcommand{\arraystretch}{1.06}
\begin{tabular}{lc}
\toprule
Variant & Top-1(\%)\\
\midrule
2D Full  & \best{72.82} \\
2D A1 (w/o Hand Joints)  & 63.68 \\
3D Full  & \best{77.61} \\
3D A1 (w/o Hand Joints) & 74.63 \\
\bottomrule
\end{tabular}
\vspace{-0.5cm}
\end{table}

As reported in Table~\ref{tab:target}, the full 2D and 3D models achieve 72.82\% and 77.61\% Top-1 target-ranking accuracy, respectively. Removing articulated-joint input reduces accuracy by 9.14 and 2.98 percentage points. 
These results characterize the H2O-PT interaction head trained with auxiliary target and verb supervision. They complement the trajectory-based ablations A5 and A8.

\subsection{Learned Operation Representations and Candidate Memory}
\label{sec:representation}
\begin{figure}[!t]
\centering

\subfloat[Stage-token PCA for two sequences, colored by predicted latent operation components.
\label{fig:stage_pca}]{%
    \includegraphics[width=0.75\columnwidth]{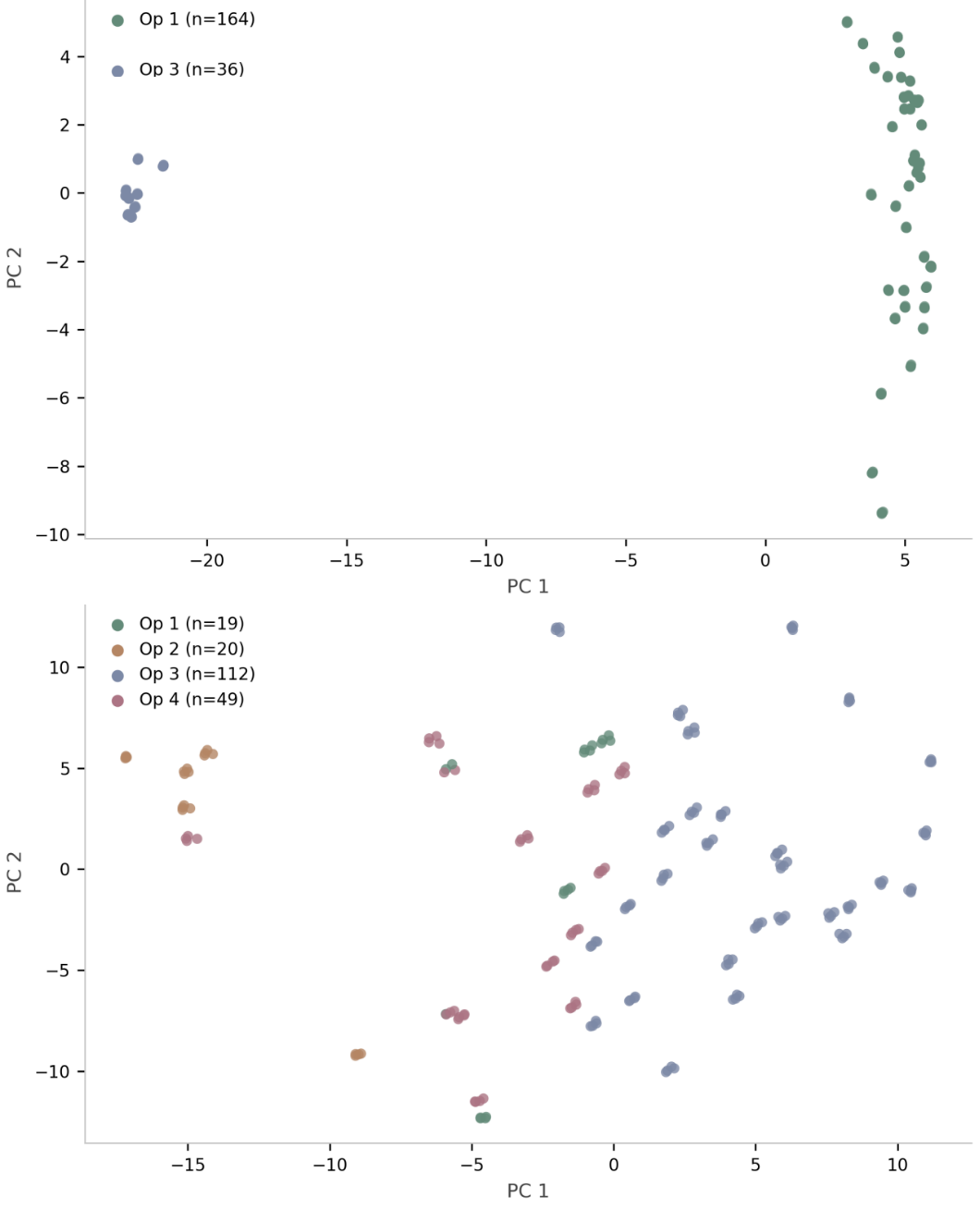}%
}
\vspace{-0.5cm}

\par\medskip

\subfloat[Candidate-interaction PCA with $K$-means clusters. The bread-associated region contains multiple latent operation components.
\label{fig:candidate_pca}]{%
    \includegraphics[width=0.75\columnwidth]{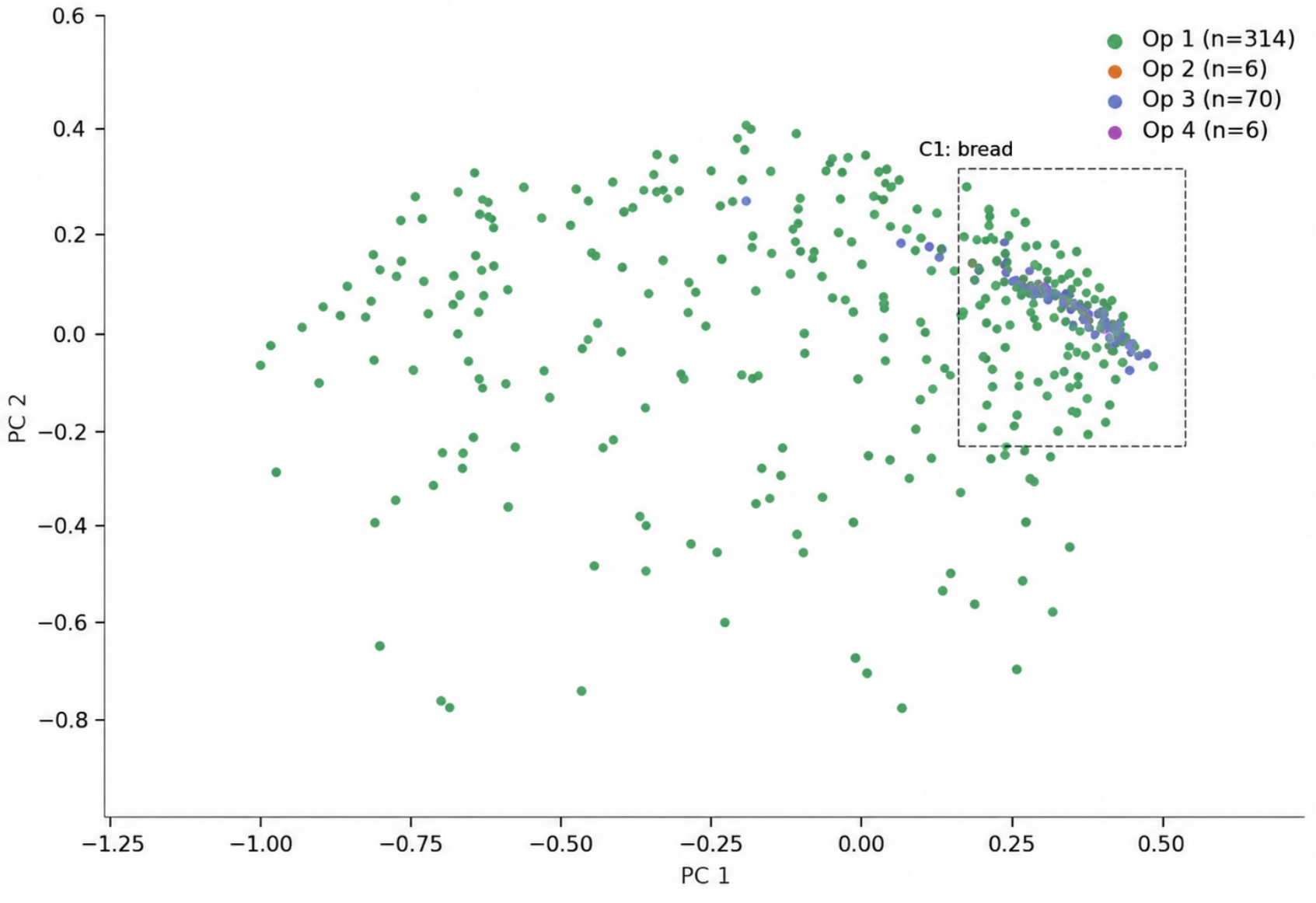}%
}

\caption{Visualization of learned operation and candidate-interaction representations. PCA projects high-dimensional features into two dimensions. Candidate names are inferred from CLIP, and colors indicate dominant operation components in the learned representation.}
\label{fig:operation_pca}
\vspace{-0.5cm}
\end{figure}

The learned interaction space is examined at two levels in Fig.~\ref{fig:operation_pca}. The stage-token projection displays different patterns of latent operation occupancy across sequences, while the candidate-interaction projection relates those patterns to the objects that may be manipulated.

In Fig.~\ref{fig:operation_pca}(a), PCA maps high-dimensional stage tokens onto their first two principal components, with colors indicating the operation component predicted by the model. The upper example occupies two components, while the lower example occupies all four. The distinct distributions indicate that stage representations capture different operation patterns across sequences. These axes are feature-space directions, not spatial coordinates or time.

Figure~\ref{fig:operation_pca}(b) visualizes candidate-interaction tokens using PCA, with dashed boxes indicating $K$-means clusters. Colors represent learned operation components rather than semantic action labels. The highlighted $\mathrm{C1{:}\ bread}$ region contains multiple components, illustrating that different latent operations can be associated with the same object.

\begin{figure}[!t]
\centering
\includegraphics[width=\columnwidth]{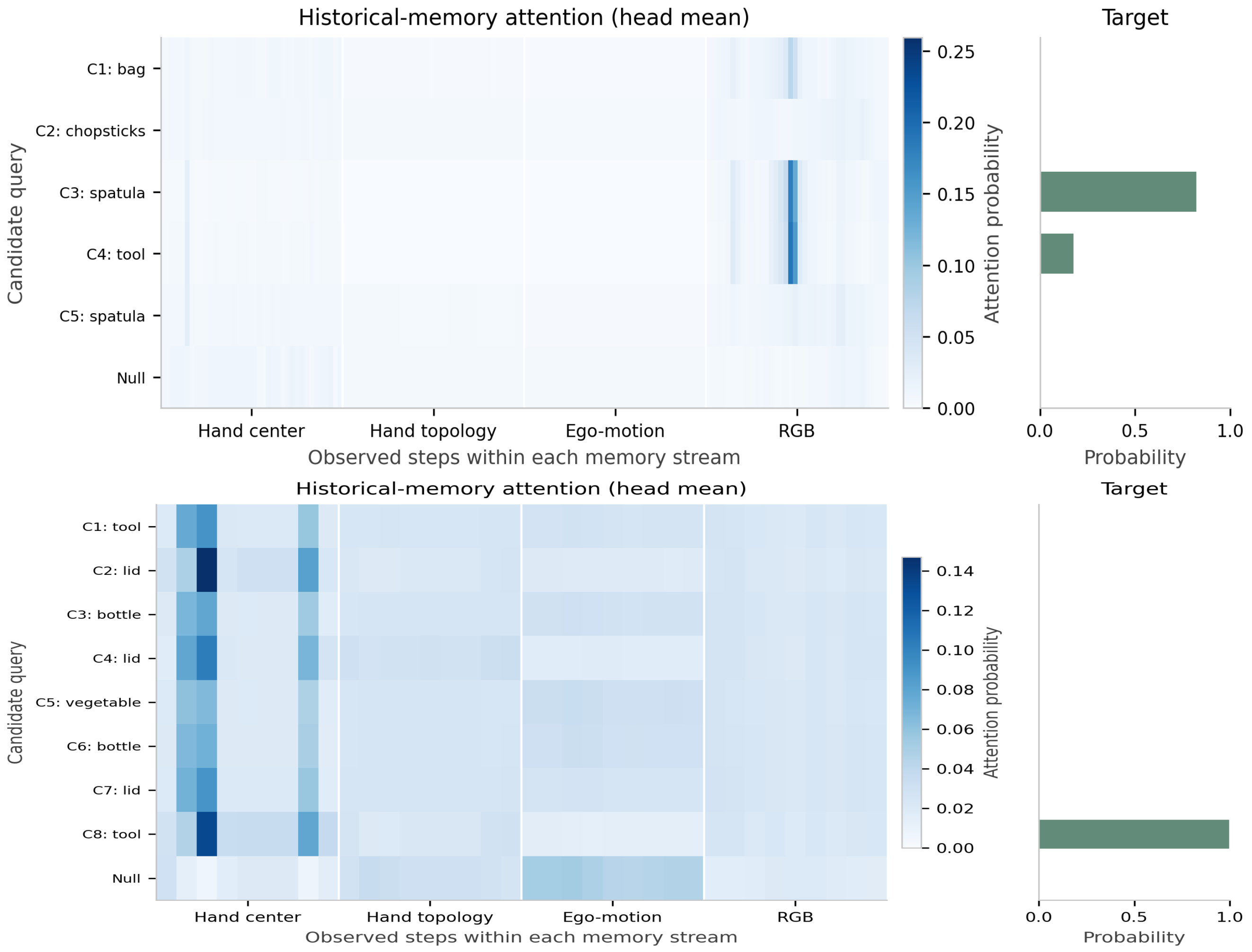}
\vspace{-0.75cm}
\caption{Candidate-conditioned historical evidence retrieval. The heat maps show final-layer, head-averaged attention from valid candidate queries and null to hand-center, hand-topology, egomotion, and RGB memory. Bars report native object-selection probabilities. Candidate names are inferred from CLIP.}
\label{fig:memory_attention}
\vspace{-0.5cm}
\end{figure}

Candidate-aware retrieval further reveals how the learned interaction representations incorporate historical evidence. Figure~\ref{fig:memory_attention} shows last-layer, head-averaged candidate queries over the observed hand-center, hand-topology, egomotion, and RGB memory streams, alongside native target probabilities.

In the upper example, candidate C3 receives approximately $0.823$ target probability and retrieves most of its historical information from RGB ($80.9\%$) and hand-center tokens ($14.8\%$). In the lower example, candidate C8 receives approximately $0.997$ probability while drawing on all four streams. These candidate-specific retrieval patterns demonstrate how the model adapts its use of observed evidence to the potential interaction target. Together with the target-ranking results, they connect the learned hand--object representation to its object selection behavior.

\subsection{Ordered Stage Dynamics and Geometric Refinement}
\label{sec:temporal_analysis}

\begin{figure}[!t]
\centering
\includegraphics[width=\columnwidth]{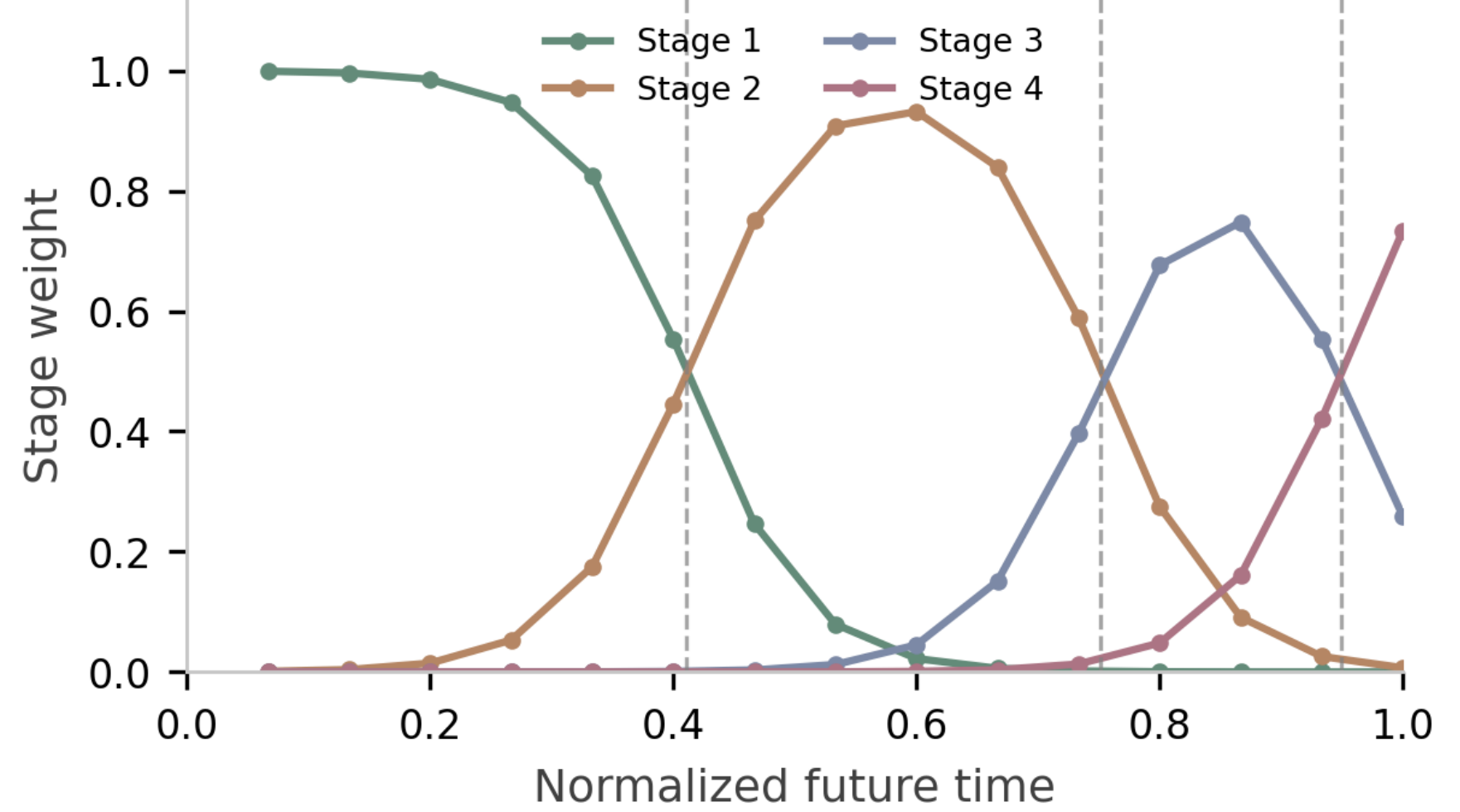}
\vspace{-0.75cm}
\caption{Learned ordered motion stages. Stage weights vary over normalized future time, while dashed vertical lines indicate predicted transition positions.}
\label{fig:temporal_stages}
\vspace{-0.5cm}
\end{figure}

\textbf{Ordered temporal stages.}
Figure~\ref{fig:temporal_stages} visualizes the temporal organization learned by TOMAS. Each stage contributes a normalized weight to the future motion representation, and neighboring stages overlap around their inferred transitions.

The stage weights progress smoothly, with transition positions near $0.411$, $0.752$, and $0.949$ of the prediction horizon. Their unequal temporal spans illustrate how the model represents nonuniform motion progression. The A6 ablation evaluates the contribution of the ordered-boundary parameterization relative to unconstrained softmax temporal attention.

\begin{figure}[!t]
\centering
\includegraphics[width=\columnwidth]{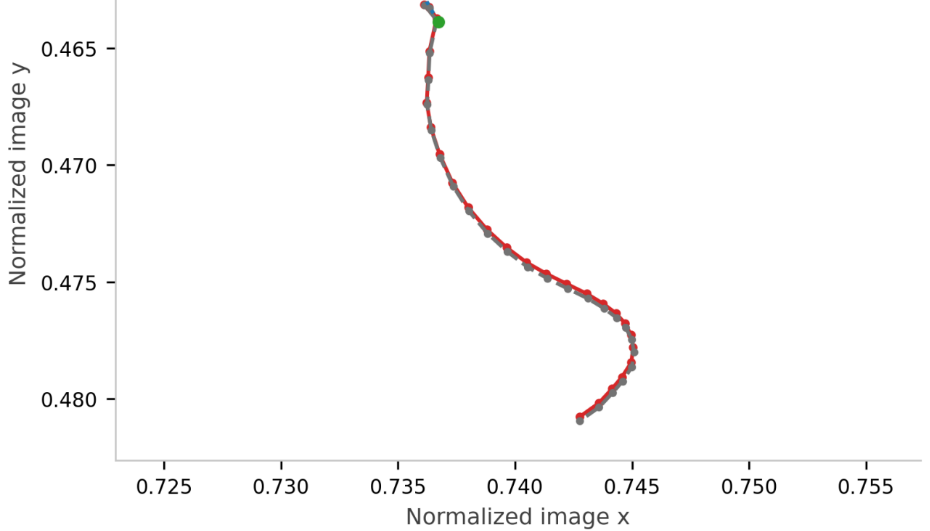}
\vspace{-0.75cm}
\caption{Local effect of subgoal-guided refinement. The full prediction is shown in red, and the prediction without final-pass relative-subgoal-coordinate input is gray and dashed. Green indicates the last observed hand position.}
\label{fig:geom_refinement}

\vspace{-0.5cm}
\end{figure}

\textbf{Geometric trajectory refinement.}
Figure~\ref{fig:geom_refinement} examines how candidate-relative subgoal coordinates influence the predicted path. The model weights and the maximum a posteriori (MAP) interaction hypothesis remain fixed; only the relative-subgoal-coordinate input in the final refinement pass is disabled.

For the illustrated sample, the full model reduces ADE/FDE from $1.805/4.032$ to $1.685/3.755$ in the original diagnostic units, corresponding to relative reductions of $6.66\%$ and $6.88\%$. The resulting path illustrates how geometric conditioning adjusts the trajectory beyond its temporal stage allocation. This local intervention complements the dataset-level A7 ablation on geometric subgoals.

\section{Conclusion}
TOMAS introduces a topology-aware interaction-to-motion forecasting framework that conditions future trajectory generation at both global and stage-wise levels. Candidate-specific latent operation hypotheses organize ordered temporal stages and object-relative geometric refinement, linking interaction reasoning to the development of continuous hand motion. Across seven evaluated 2D/3D configurations, TOMAS achieves the lowest ADE, and targeted ablations show the complementary value of global interaction conditioning, stage-level operation conditioning, and ordered stage boundaries. Native target-ranking results further demonstrate that the interaction branch can identify plausible object candidates from observed history. This coupling of interaction hypotheses with stage-structured motion offers a practical basis for more expressive egocentric forecasting.
\bibliographystyle{IEEEtran}
\bibliography{references}

% =============================================
% Author Biographies
% =============================================

\begin{IEEEbiography}[{\biophoto{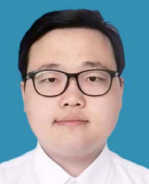}}]{Guanzhong Sun}
received the B.Eng. degree in mechanical engineering from Nantong University in 2018, and the M.Eng. degree in mechatronic engineering in 2022 from the China University of Mining and Technology, where he is currently working toward the Ph.D. degree in artificial intelligence. His current research interests include robotic manipulation.
\end{IEEEbiography}

\biospace
\begin{IEEEbiography}[{\biophoto{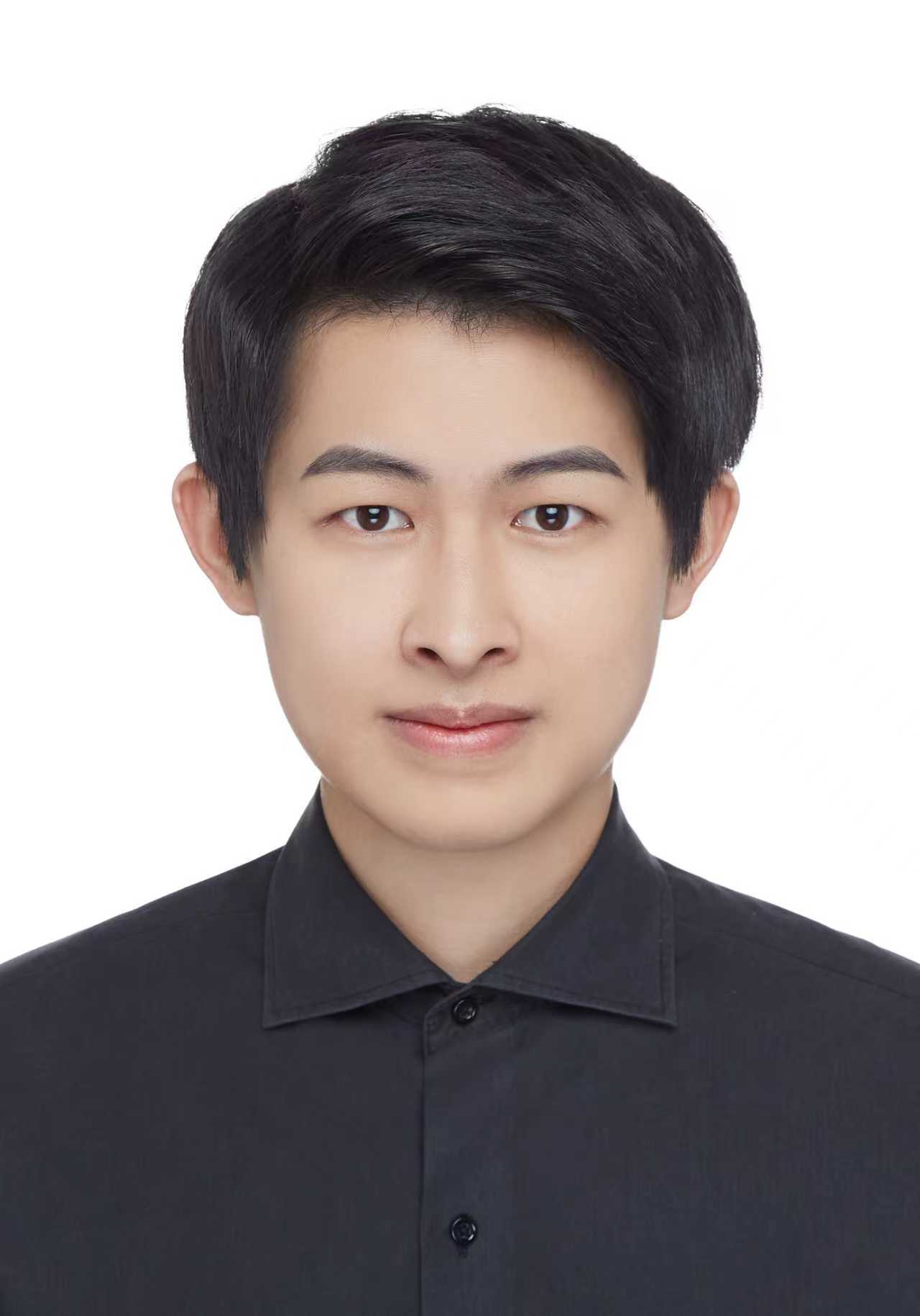}}]{Junyi Ma}
is a Ph.D. candidate at IRMV Lab, Shanghai Jiao Tong University. He received his Master's degree and Bachelor's degree from the Beijing Institute of Technology, in 2020 and 2023 respectively. He is currently supervised by Prof. Hesheng Wang. His research interests include human motion analysis, hand-object interaction, egocentric vision, and human-robot skill transfer.
\end{IEEEbiography}

\biospace
\begin{IEEEbiography}[{\biophoto{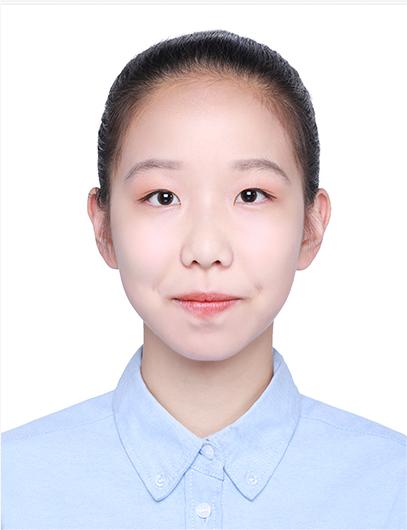}}]{Yuxuan Wu}
is a Ph.D. candidate at IRMV Lab, Shanghai Jiao Tong University. She received her Bachelor's degree from Beihang University in 2024. She is currently supervised by Prof. Hesheng Wang. Her research interests include knowledge transfer for robotic manipulation and continual learning of robotic manipulation policies.
\end{IEEEbiography}

\biospace
\begin{IEEEbiography}[{\biophoto{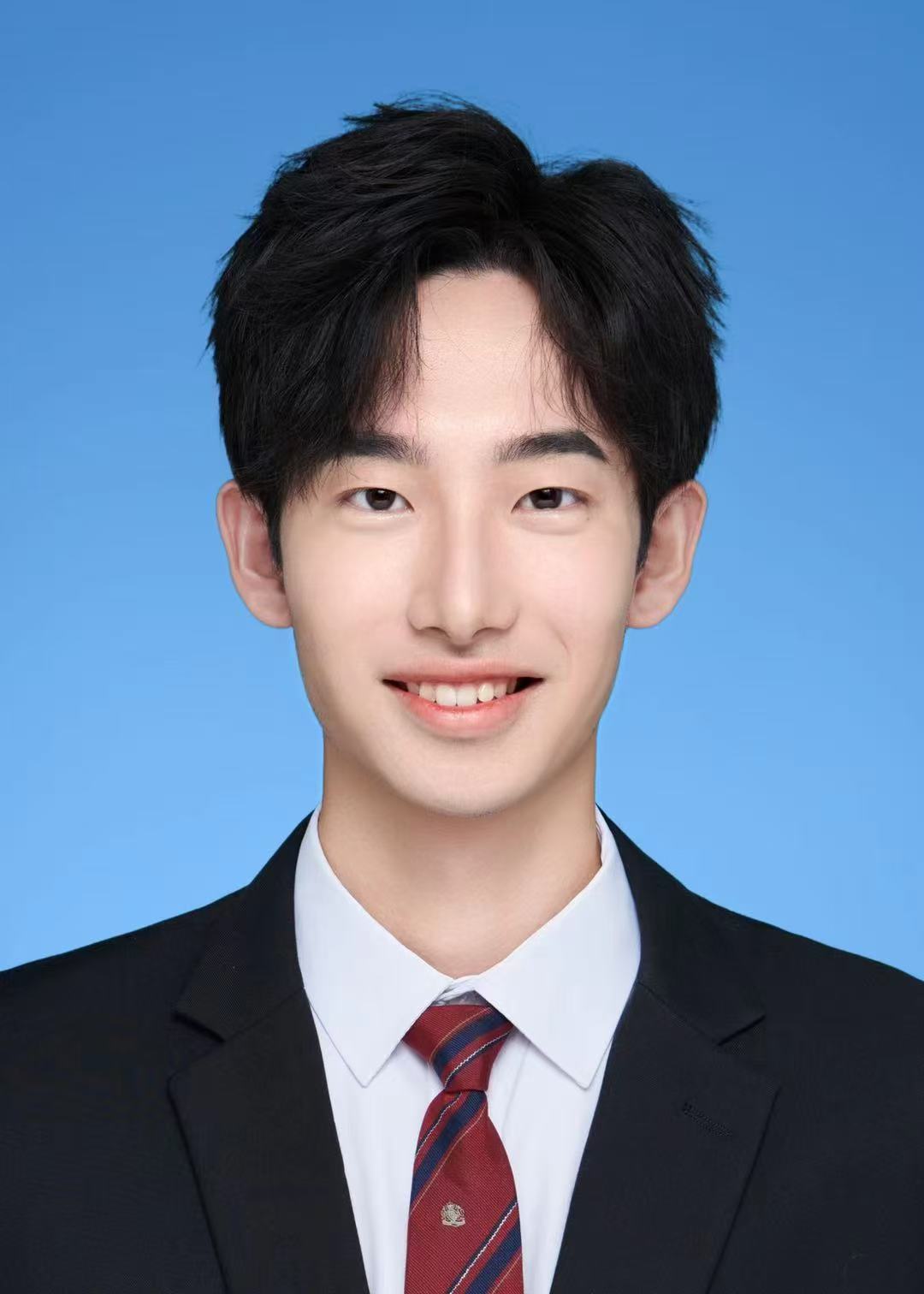}}]{Erhang Zhang}
is a Ph.D. candidate at IRMV Lab, Shanghai Jiao Tong University. He received his Bachelor's degree from Shandong University in 2025. He is currently supervised by Prof. Hesheng Wang. His research interests include large-scale foundation models, human-robot skill transfer and tactile representation learning.
\end{IEEEbiography}

\biospace
\begin{IEEEbiography}[{\biophoto{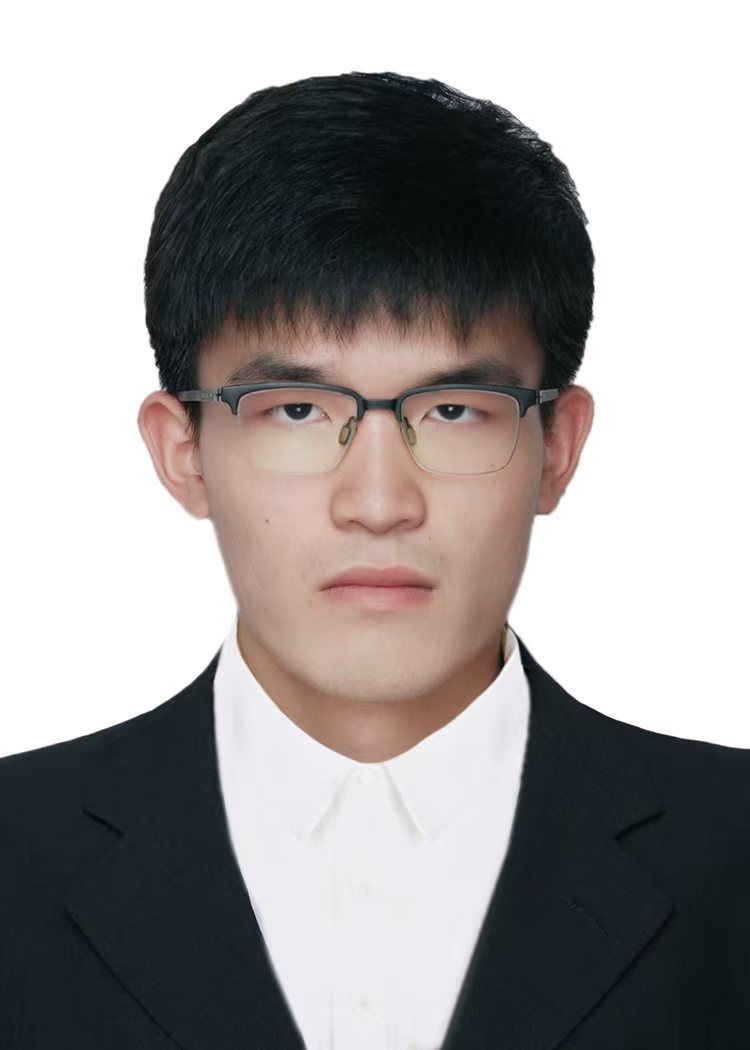}}]{Chenyang Xu}
is a Ph.D. student at IRMV Lab, Shanghai Jiao Tong University. He received his Bachelor's degree from Huazhong University of Science and Technology in 2025. He is currently supervised by Prof. Hesheng Wang. His research interests include robot learning, mobile manipulation, and human-robot interaction.
\end{IEEEbiography}

\biospace
\begin{IEEEbiography}[{\biophoto{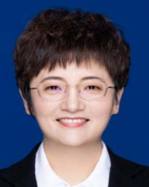}}]{Yanzi Miao}
(Member, IEEE) received the Ph.D. degree in control science and engineering from the China University of Mining and Technology, Xuzhou, China, in 2009. As a joint-Ph.D. candidate and a visiting scholar, she was with the Department of Informatics, University of Hamburg, Germany, in 2007 and 2017, respectively. 
She is currently a Professor with the School of Information and Control Engineering, China University of Mining and Technology. Her current research interests include intelligent perception and fusion, machine vision, and robots in mine. Dr. Miao was the Technical Co-Chair of the IEEE International Conference on Real-time Computing and Robotics in 2019.
\end{IEEEbiography}

\biospace
\begin{IEEEbiography}[{\biophoto{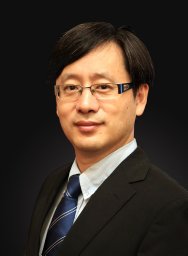}}]{Hesheng Wang}
(Senior Member, IEEE) received the B.Eng. degree in electrical engineering from the Harbin Institute of Technology, Harbin, China, in 2002, and the M.Phil. and Ph.D. degrees in automation and computer-aided engineering from The Chinese University of Hong Kong, Hong Kong, in 2004 and 2007, respectively. He is currently a Distinguished Professor with the School of Automation and Intelligent Sensing, Shanghai Jiao Tong University, Shanghai, China. His current research interests include visual servoing, intelligent robotics, computer vision, and autonomous driving.

Dr. Wang is an Associate Editor of Robotic Intelligence and Automation and the International Journal of Humanoid Robotics, a Senior Editor of the IEEE/ASME Transactions on Mechatronics, and an Editor-in-Chief of Robot Learning. He served as an Associate Editor of the IEEE Transactions on Robotics from 2015 to 2019 and the IEEE Transactions on Automation Science and Engineering from 2021 to 2023. He was the General Chair of IEEE/RSJ IROS 2025, IEEE ROBIO 2022, and IEEE RCAR 2016.
\end{IEEEbiography}

\end{document}